\pdfoutput=1

\documentclass[11pt]{article}
\usepackage[table,usenames,dvipsnames]{xcolor}
\usepackage{booktabs}
\usepackage{graphicx}
\usepackage{svg}
\usepackage{hyperref}
\usepackage{acl}

\definecolor{lavender}{rgb}{0.9, 0.9, 0.98}
\definecolor{redv}{HTML}{EF767A}
\definecolor{bluev}{HTML}{427AA1}
\definecolor{greenv}{HTML}{2BA84A}

\definecolor{a}{HTML}{EF767A}
\definecolor{b}{HTML}{427AA1}
\definecolor{c}{HTML}{2BA84A}
\definecolor{d}{HTML}{2BA84A}

\newcolumntype{b}{>{\columncolor{lavender}}c}

\usepackage{times}
\usepackage{latexsym}

\usepackage[T1]{fontenc}

\usepackage[utf8]{inputenc}

\usepackage{microtype}

\title{A Copy Mechanism for Handling Knowledge Base Elements in SPARQL Neural Machine Translation}

\author{Rose Hirigoyen, Amal Zouaq \and Samuel Reyd\\
  \texttt{rose.hirigoyen@polymtl.ca}\\ \texttt{amal.zouaq@polymtl.ca} \\\texttt{samuel.reyd@polymtl.ca}  \\
    Polytechnique Montreal
  }

\begin{document}
\maketitle
\begin{abstract}
Neural Machine Translation (NMT) models from English to SPARQL are a promising development for SPARQL query generation. However, current architectures are unable to integrate the knowledge base (KB) schema and handle questions on knowledge resources, classes, and properties unseen during training, rendering them unusable outside the scope of topics covered in the training set.  Inspired by the performance gains in natural language processing tasks, we propose to integrate a copy mechanism for neural SPARQL query generation as a way to tackle this issue. We illustrate our proposal by adding a copy layer and a dynamic knowledge base vocabulary to two Seq2Seq architectures (CNNs and Transformers). This layer makes the models copy KB elements directly from the questions, instead of generating them. We evaluate our approach on state-of-the-art datasets, including datasets referencing unknown KB elements and measure the accuracy of the copy-augmented architectures. Our results show a considerable increase in performance on all datasets compared to non-copy architectures.
\end{abstract}

\section{Introduction}

The Semantic Web organizes concepts in optimized, machine-readable, 
knowledge bases (KB) (or knowledge graphs). Still, as these knowledge bases are not immediately designed with a human user in mind, the SPARQL Protocol and RDF Query Language (SPARQL) is hardly accessible to laypeople with little-to-no knowledge of programming languages. This creates a strong accessibility bias, as it prevents users from accessing sizeable amounts of information because of their lack of a specific skillset.

One way to bypass any need for prior knowledge is by allowing the users to query KBs using natural language questions. Figure~\ref{fig:taskexample} illustrates the task at hand. 
More specifically, using neural machine translation (NMT) to translate natural language questions to SPARQL queries has proven to be an interesting avenue to solve this challenge, with BLEU-score performances of more than 90\% across multiple datasets~\cite{TNTSPA}.

\begin{figure}
\begin{center}
\begin{tabular}{p{75mm}}
\toprule
\textbf{Q: }What is Villa La Mauresque ?\\
\midrule
\rowcolor{lavender}
\texttt{select ?a where}\\
\rowcolor{lavender}
\texttt{\hspace*{1cm}\{ dbr:Villa\_La\_Mauresque}\\
\rowcolor{lavender}
\texttt{\hspace*{1cm}dbo:abstract ?a \}}\\
\midrule
\textbf{A: }The villa La Mauresque is located in cap Ferrat (Alpes-Maritimes) and was remodeled in 1927 ... \\
\bottomrule
\end{tabular}
\end{center}
\caption{Example of the SPARQL NMT task}
\label{fig:taskexample}
\end{figure}

However, behind these high-performing architectures are models that rarely return the correct answer to a question about a topic they have never seen in training, even if the information is available in the KB. 
As a single wrong answer can negatively affect the user's trust in the model, this limitation becomes a critical downfall for an automatic SPARQL query generation model. 
The main goal of this paper is to propose a mechanism to effectively generate accurate SPARQL queries. In particular, we aim at handling out-of-vocabulary (OOV) knowledge base elements at the schema level ( classes, properties) and the instance level. As such, we put forth the following research questions:
\begin{itemize}
    \item \textbf{RQ1: } Is the integration of the KB elements in the question sufficient for the model to handle OOV elements?
    \item \textbf{RQ2: } Is the accuracy of the translations improved if the neural translation architecture is able to copy KB elements directly from the question?
    \item \textbf{RQ3: } Does the evaluation of the model on a dataset composed solely of unknown KB elements allows for a complete overview of the model's capabilities?
\end{itemize}

Our main contributions are as follows. (1) Given a working tagging algorithm, we propose a way to allow NMT models to handle questions on topics they have not seen during training. (2) We propose a methodology to evaluate a model's performance exhaustively. (3) Finally, we produce standardized, corrected, and tagged versions of the datasets to foster reproducibility and future developments in this research field\footnote{\url{https://github.com/Lama-West/SPARQL\_Query\_Generation\_aacl-ijcnl2022}}.

\section{Related Work}

\paragraph{Knowledge Bases Terminology.} 
A \textit{knowledge base (KB)} stores data in the form of one or more Resource Description Framework (RDF) graphs, in which the nodes are concepts or instances, and the edges encode the relationship between them. 
An RDF graph is described using (subject, property, object) triples, which we refer to as \textit{KB elements}. Each KB element has a unique URI, which is used to reference it in a SPARQL query and a label, which is their name in a natural language. If there is no label, we can generate one from the element's URI.

\paragraph{Seq2Seq for NMT.}
The base architecture behind many NMT models is \textit{Seq2Seq}, which 
learns to generate words using source and target vocabularies. 
If there is a token in the source sentence that is not in the vocabulary, the model simply replaces it by the \textbf{<unk>} placeholder token. The model is as such only able to generate tokens that are in its target vocabulary. 
The transformers~\cite{Transformer} and convolutional networks~\cite{CNNS2S} are currently the two best non-pretrained architectures for SPARQL NMT, as reported by~\citet{TNTSPA}. 

As more data becomes available, an important development in this field is the introduction of pretrained language  models and their application for neural machine translation. 
For example, T5~\cite{T5} uses Transformers and transfer learning to translate three languages at once. This provides the model with a rich vocabulary of about 32000 tokens, and it can use its prior knowledge to reach higher performances on languages for which there is less training data. However, as stated in the paper, the model can only process a predetermined, fixed set of languages and it uses a fixed vocabulary. This means that as much as it is able to infer information in general translation problems, it encounters the same OOV problem as other Seq2Seq-type models, since it does not have the ability to learn new 
words once training is over. 
Very recent concurrent efforts explore the use of pretrained language models for SPARQL query generation \cite{SPARQLParsing}.
For example, SGPT \cite{SGPT} is built on GPT-2 \cite{GPT2} and aims to generate SPARQL queries by encoding linguistic features of questions and the knowledge graph. 
It uses an entity masking strategy 
and generates queries with placeholders. After a query is generated by the neural architecture, a \textit{post-processor} places the correct KB elements in the right places in the query. While our objective is similar, our approach aims at using a copy mechanism directly in the Seq2Seq architecture to place KB elements in the question instead of doing it in a post-processing step.

\paragraph{KGQA Systems.}
Since the handling of OOV KB elements is limited in the specific field of SPARQL NMT, it is necessary to broaden our research and learn from similar SPARQL NLP tasks. In particular, Knowledge Graph Question Answering (KGQA) systems aim to reconstruct a subgraph of the RDF schema from a natural language question and use it to generate a correct query. A notable aspect of these architectures is that  they can provide a correct answer to a question on a topic not seen in training~\cite{KGQASurvey} (if the answer is in the KB). 
An interesting KGQA system is HGNet~\cite{HGNet}. 
A key aspect of this architecture is that in trying to generate the subgraph necessary to answer the question, it can take advantage of the fact that such graph often contain duplicated vertices. It uses LSTMs and a copy mechanism to copy these duplicated vertices, thus facilitating the generation task. Such systems ~\cite{HGNet, TEBAQ} highlight the importance of integrating the RDF schema and resources in the architecture. Doing so not only provides us with additional information on the KB elements themselves, but also on the elements which they are related to and which are more likely to be referenced as well.

\paragraph{SQL Systems.}
It is also useful to explore what we can learn from problems similar to the one of SPARQL NMT, such as the text-to-SQL semantic parsing problem \cite{RatSQL, DuoRat}. 
One of the current best performing model  \cite{SQLBest} is not a Seq2Seq-type model, but rather a classification model that learns to predict 6 different SQL components by leveraging the extensively annotated WikiSQL dataset. 
Seq2SQL ~\cite{Seq2SQL} is another approach, which, while not the best performing architecture,  is worth noting for its schema integration mechanism. 
Seq2SQL augments the natural language question by concatenating it to all the columns' names and to the SQL vocabulary. The schema is essentially integrated directly in the input. Once again, incorporating the schema in the architecture gives the model enough information to understand which database elements (or for SPARQL, KB elements) are referenced in a question whether or not it has seen them during training, provided they are available in the database.

\paragraph{Copy Mechanism.}
The copy mechanism has shown its effectiveness in several encoder-decoder NLP tasks such as summarization~\cite{summarization}, grammatical errors correction~\cite{gram}, and knowledge graph question answering (KGQA)~\cite{HGNet}. However, to our knowledge, it has not yet been used in SPARQL NMT as we propose here.
Our hypothesis is that, given a working tagging algorithm where, in the NL question, mentions related to a KB element are replaced by their KB URI, a model could learn to copy the KB URIs from the question to the query instead of generating them. 
Notably, we propose to integrate CopyNet \cite{CopyNet}, whose copy mechanism comes after the decoder. For each token of the output sentence, it uses attention to calculate the probability that the token should be generated from the target vocabulary, and the probability that the token should be copied directly from the source. The chances of copying are slightly higher for OOV words in the source sentence.  

\paragraph{Limitations.}

As reported by~\citet{TNTSPA}, the current best performing non-pretrained architectures  for SPARQL NMT are the \textit{Transformer Seq2Seq} and the \textit{ConvSeq2Seq}, which are Seq2Seq-type models where the encoder and decoder are respectively transformers and convolutional networks. As such, they encounter the same limitation as all Seq2Seq-type models, which is that because of the use of fixed vocabularies, the models are unable to fully handle OOV tokens. In SPARQL NMT systems, this results in the models not being able to answer questions referencing KB elements that were unseen during training. Instead, when encountering a question on a new KB element, the models generate a query referencing the element seen the most in the current context, even if it is not the one referenced in the question.

This also means that the model might learn the meaning of a specific KB element during training, but never use this knowledge if the element is not referenced in the test set.
In the context of a query language, our hypothesis is that the encoder-decoder model should focus on learning the syntax of the correct SPARQL query related to a question, instead of trying to learn the meaning of each KB element. Keeping in mind that the prevalent KBs such as DBpedia can contain tens of thousands of different URIs, expecting the model to learn everything from examples is not optimal. Furthermore, the lack of real-world data is the field of SPARQL NMT makes this approach unrealistic.

In light of these limitations, the impressive BLEU-scores  reported by~\citet{TNTSPA} raise some questions on the ability of these metrics and current datasets to properly evaluate NMT SPARQL models. Knowing that the models are only able to generate tokens learned during training, it is almost impossible for them to return a correct answer on a question whose topic is unknown, except by accident or when the expected answer is empty. Some datasets contain a number of queries that return empty answers. As such, it is important to make sure that models are thoroughly tested, especially on questions mentioning KB elements never seen during training.


\section{Architectures}
\subsection{Base Architectures}
This section describes the two best non-pretrained architectures for SPARQL NMT as reported by \citet{TNTSPA}, as well as our contribution. 

\paragraph{ConvS2S.} The convolutional sequence to sequence model (ConvS2S) is a Seq2Seq-type model where the encoder and decoder are convolutional networks \cite{CNNS2S}. 
Both the encoder and the decoder generate token embeddings and position embeddings of the vectors they receive as input, respectively the encoding of the question and the encoding of the query. The decoder also receives the output of the encoder as input, and its input vector is padded at the beginning. This creates an offset which allows the model to learn from previous words and not from the current words which it is supposed to predict. Then, the sum of the token and position embedding vectors passes multiple times through a recurrent layer. This layer comprises a 1-dimension convolution and a Gated Linear Unit (GLU) in the encoder, followed by multi-head attention in the decoder.
Following the survey by \citet{TNTSPA}, we use the same architecture configuration as FairSeq's \textit{fconv\_wmt\_en\_de} NMT architecture \cite{FairSeq}, described in Table \ref{tab:config}.
\begin{table}[htbp]
\centering
\begin{tabular}{lcp{21mm}}
\toprule
\textbf{Model} & \textbf{Transformer} & \textbf{ConS2S} \\
\midrule
\rowcolor{lavender}
Batch Size & 128 & 128 \\
Layers & 6 & 15 \\
\rowcolor{lavender}
Hid. Dim. & 1024 & [(512, 3) * 9, (1024,3) * 4, (2048, 1) * 2] \\
Dropout &  0.5 & 0.2 \\
\rowcolor{lavender}
LR & 0.0005 & 0.5\footnotemark \\
Optimizer & Adam & SGD \\
\bottomrule
\end{tabular}\\
\caption{Configuration of our Architectures}\label{tab:config}
\end{table}
\footnotetext{For the dataset TNTSPA, we used a LR of 3.5}

\paragraph{Transformer.}
The Transformer model is a Seq2Seq-type model where the encoder and decoder are transformers \cite{Transformer}. The encoder and decoder receive the same inputs as the ConvS2S. The decoder uses a multi-head attention layer that is not in the encoder. Our implementation is based on the FairSeq implementation~\cite{FairSeq} of the \textit{transformer\_iwslt\_de\_en} architecture, as described in Table \ref{tab:config}.
\subsection{A Copy-augmented Architecture} 
\begin{figure}
    \centering
    \includegraphics[width=8cm]{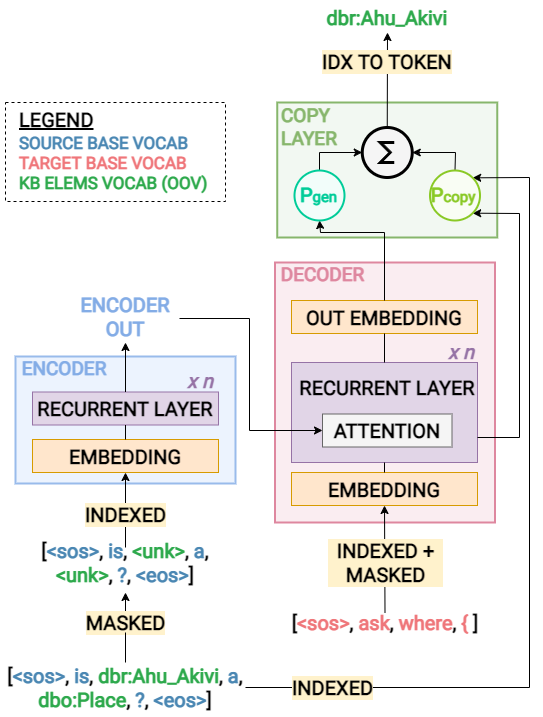}
    \caption{Encoder and copy-augmented decoder structure and interaction}
    \label{fig:arch}
\end{figure}

Figure \ref{fig:arch} shows our generic architecture, which enriches any encoder-decoder model (e.g. CNNs or transformers) with a copy layer in the decoder. It generates specific source and target vocabularies that include the KB elements as explained below.

\paragraph{Vocabularies.} 
In the baseline architectures (without copy), the source vocabulary comprises every token of the questions, and the target vocabulary comprises every token in the queries. Tokens are added in the order in which they are encountered.

However, when using the copy layer, there needs to be a way to differentiate tokens that are part of the base vocabularies (which the model will learn to generate) and tokens that are KB elements (which the model will learn to copy from the source). The latter are identifiable by their prefix, meaning tokens that start with \textbf{dbo:}, \textbf{dbr:}, \textbf{dbp:}, \textbf{dbc:}, \textbf{geo:}, \textbf{georss:} or \textbf{dct:}.  Also, since the model receives vectors of indices and not words, tokens copied from the source to the target sentence must have the same index in both the source and target vocabularies.

To accommodate these constraints, we create a base source vocabulary and a base target vocabulary containing all tokens in the inputs but no KB elements and pad them with filler words so they are the same size. 
Then, we extract the KB elements in a vocabulary extension that contains all elements in both the questions and the queries. Finally, the KB vocabulary is concatenated to each base vocabulary to create our source and target vocabularies. 

As we know the cutoff index of the initial vocabularies, we can quickly determine that each index above this cutoff represents a KB element we want to copy. During inference, if a new KB element is encountered, we can add it at the end of our source and target vocabularies, giving the model the capacity to copy it.

\paragraph{Copy Layer.}
In a copy-augmented architecture, the encoder and decoder receive masked source and target vectors, meaning any token above the cutoff index (and as such, out of the vocabulary) is replaced by a 0, representing an unknown token. As the role of the copy layer is to handle KB elements, this masking lets the encoder and decoder focus on the syntax rather than on the KB elements.

The copy layer comes after the decoder. It takes as input the unmasked encoded question and the decoder output, comprised of the attention scores and the probability of generating each word of the base target vocabulary. Ported to the Transformer architecture by \cite{summarization, gram}, we were able to adapt it to ConvS2S since both generate multi-head attention scores.

First, we identify whether there are any KB elements amongst the tokens of the encoded question by using the cutoff index. If it is the case, we extend the output probability tensor to include these extra tokens and initially assign them a generation probability of 0. Then, we calculate the probability of each token being generated, which is the softmax of the probability tensor. Using the attention score, we also calculate the probability of each word being copied directly from the source sentence. Following the implementation of \cite{gram}, we compute a balancing factor $\alpha_{bal} \epsilon [0,1]$ between the copy and the generation probabilities using Equation \ref{eq:balancing}, where Q, K and V are the query, key and value needed to calculate attention and $W\textsuperscript{T}$ is a learnable parameter. The final probability of each token being the next word is the sum of the generation and copy probabilities balanced by this factor.

\begin{equation}
A_t = Q^T * K
\end{equation}
\begin{equation}\label{eq:balancing}
\alpha_{bal} = sigmoid(W^T*(A_t^T*V))
\end{equation}

\section{Methodology}
\subsection{Datasets}

\paragraph{Format.} Most natural language (NL) to SPARQL datasets are generated using templates to compensate for the lack of real-world data. A template is an NL question and its corresponding SPARQL query, in which there are annotated blanks to indicate the types of the KB element to insert (resources, classes, properties). These blanks are then replaced by KB elements' labels in the questions, and KB URIs in the queries. Many datasets also use an alternate version of SPARQL introduced by~\cite{NSPM} called \textit{intermediary SPARQL}, in which each symbol (e.g., brackets, dots) is replaced by a specific natural language expression. This encoding aims to make SPARQL closer to a natural human language. URIs are also reduced using their prefixes. To return to the original executable SPARQL query, one only has to make the inverse permutations. 
Table \ref{tab:datasets} shows the datasets used in this work. We split the datasets in an 80-10-10 fashion to reproduce the results reported by \cite{TNTSPA}.

\begin{table}[htbp]
\centering
\begin{tabular}{lccc}
\toprule
 & \textbf{Mon} & \textbf{Mon50} & \textbf{Mon80}\\
\midrule
\rowcolor{lavender}
Train & 1797 & 1787 & 1791\\
Test & 815 & 825 & 816\\
\rowcolor{lavender}
Int. rate & 0.928 & 0.925 & 0.925 \\
\toprule
& \textbf{TNTSPA} & \textbf{LC-QuAD} & \textbf{DBNQA} \\
\midrule
\rowcolor{lavender}
Train & 4153 & 4150 & 145 429 \\
Test & 1045 & 1066 & 38 348\\
\rowcolor{lavender}
Int. rate & 0.704 & 0.713 & 0.797 \\
\bottomrule
\end{tabular}
\caption{Summary of the distribution of KB elements in the datasets}\label{tab:datasets}
\end{table}

\paragraph{Monument.}
The Monument dataset ~\cite{NSPM} consists of pairs of English natural questions and intermediary SPARQL queries generated from 38 templates.  
The authors ~\cite{TNTSPA} generate other versions of the dataset: Monument, Monument50 and Monument80. The three versions are very similar in that they are all generated using 600 examples per template with different combinations of KB elements. 
We used their versions to be able to compare our results to state-of-the-art architectures. The high BLEU scores reported by \citet{TNTSPA} are explained by the fact that most KB elements in the test set have already been seen during training, as shown by the high intersection rate in Table \ref{tab:datasets}. Also, this dataset covers fewer KB elements in more entries, which gives the models plenty of examples to learn each element in its context. Overall, good results on this dataset only mean a model is functional.

\paragraph{LC-QuAD.} The LC-QuAD datasets provide entries of multiple types (COUNT, ASK, SELECT) and cover a broad range of KB elements. We prioritized LC-QuAD v1.0~\cite{LCQUAD} over the newer LC-QuAD v2.0~\cite{LCQUAD2} since the models to which we compare our work are trained on the first version. Further tests on LC-QuAD 2 are left for future work.

In LC-QuADv1.0, each entry contains an English natural language question and its corresponding SPARQL query generated from a template (called intermediary question), as well as a version of the question reformulated by an expert (called corrected question). It comprises 5000 entries generated from 33 of the 43 templates available. Table \ref{tab:datasets} shows that it is much more challenging than Monument. Indeed, there are many more different KB elements, fewer examples per element, and a lower intersection rate between the train and test sets. 

We use three versions of the LC-QuAD dataset. The first version, referred to as \textbf{LC-QuAD Intermediary Questions}, uses the intermediary questions and their corresponding queries. These questions use the formulations defined by the templates. The second and more challenging version, referred to as \textbf{LC-QuAD Corrected Questions}, uses the reformulated natural language questions of the dataset and their corresponding queries. The third version, referred to as \textbf{TNTSPA}, is the version generated by the authors of the survey \cite{TNTSPA}. It contains the reformulated questions (formulated in a more natural way) and queries found in the LC-QuADv1.0 dataset, but is split differently. Since no validation set is provided for the TNTSPA dataset, we use entries from LC-QuAD v1.0 that are not in the TNTSPA train or test sets. Since there are no templates associated to this dataset, we only use it to ensure we are able to reproduce state-of-the-art results with our implementation of the baselines architectures. 

\paragraph{DBNQA.}
The DBpedia Neural Question Answering (DBNQA) dataset \cite{DBNQA} is composed of 894,499 pairs of natural language questions and SPARQL queries. The entries are generated using 5165 question-query templates, constructed from entries in the LC-QuADv1.0 \cite{LCQUAD} and QALD-7 \cite{QALD7} datasets.
We used the templates provided with the dataset but we did not manage to match all entries.
We then extracted and corrected 512 templates suitable for the annotation of the questions and used the 398,284 entries corresponding to these templates. 
We also provide directly executable SPARQL queries instead of intermediate SPARQL queries.
%

\paragraph{RDF schema integration.}
As this research focuses mainly on finding a solution for the OOV problem, we developed a rudimentary tagging algorithm that leverages the templates. For each entry, we replace the KB elements labels that replace the blanks in the questions with their corresponding URIs in the query. KB elements that would be encoded as multiple tokens because of intermediary SPARQL (e.g., \textbf{[dbr\_Cenotaph\_, attr\_open, Montreal, attr\_close]}) are encoded as a single token (e.g.,  \textbf{dbr\_Cenotaph\_(Montreal)}) to reduce the vocabulary size. This dependence on templates is why we use the LC-QuAD Intermediary Questions version of LC-QuAD to train and evaluate our copy-augmented models, as it is the only version we could tag with complete accuracy. Figure \ref{fig:example} shows an entry before and after tagging.


\begin{figure}[htp]
\begin{center}
\begin{tabular}{p{75mm}}
\toprule
\textbf{Template: }what is the \textbf{\color{Aquamarine}<domain>} whose \textbf{\color{BurntOrange}<property\_1>} is \textbf{\color{CarnationPink}<resource\_1>} and \textbf{\color{Green}<property\_2>} is \textbf{\color{RoyalBlue}<resource\_2>} ?\\
\midrule
\textbf{Question: }what is the \textbf{\color{Aquamarine}formula one racer} whose \textbf{\color{BurntOrange}relatives} is \textbf{\color{CarnationPink}ralf schumacher} and \textbf{\color{Green}has child} is \textbf{\color{RoyalBlue}mick schumacher} ?\\
\midrule
\textbf{Tagged: }what is the \textbf{\color{Aquamarine}dbo:FormulaOneRacer} whose \textbf{\color{BurntOrange}dbp:relatives} is \textbf{\color{CarnationPink}dbr:Ralf\_Schumacher} and \textbf{\color{Green}dbo:child} is \textbf{\color{RoyalBlue}dbr:Mick\_Schumacher} ?\\
\bottomrule
\end{tabular}
\end{center}
\caption{A tagged question}
\label{fig:example}
\end{figure}



\paragraph{OOV Datasets.}
Finally, we generate an additional test set of 250 entries for each dataset called the \textbf{OOV Set}. First, we go through the dataset and make a list of all the referenced KB elements. Then, we use the templates to generate entries where the placeholders are replaced by KB elements that are not in the list, effectively creating a dataset in which no KB element has been seen in training.

To avoid false positives, we built our datasets so that questions would return a non-empty answer whenever possible. However, this proved to be a challenging task and our most successful attempts still contain about 70\% of empty answers (count of 0, ask that returns false, or empty sets of elements).
False positives can happen when a query returns an empty answer regardless of the KB elements referenced (e.g. an impossible question that links unrelated KB elements, or a question for which the KB does not contain an answer).


\subsection{Evaluation}
We use two main metrics to evaluate the original test sets and the oov test sets: the \textbf{BLEU-score} 
and the \textbf{answer accuracy}, which calculates the accuracy of the answers returned by the generated queries against the expected answers. 

\section{Results}
We trained and evaluated our implementation of the models using Google Colab GPUs. We compare our results to those reported by \cite{TNTSPA}, who train their model on HPC servers using the FairSeq implementations of the CnnS2S and Transformer architectures. It is important to note that they report the peak performance while we report the average of three runs. This means that we expect slightly lower performances when reproducing their results.

\paragraph{Baseline architectures on original datasets.} Table~\ref{tab:resOG} shows the results of the baseline architectures on original datasets. We clearly reproduce the performances of the survey by \citet{TNTSPA}. Even if our results for LC-QuAD are slightly lower, it is still within an acceptable margin. Because of the randomness of the weights initialization, the performance difference between a good and a under-performing run can be up to ten points. This margin also accounts for the small difference between TNTSPA and the LQ Corr Qsts. The higher scores on LQ Intrm.Qsts compared to the corrected questions are explained by the fact that the questions are generated from templates. This results in a smaller source vocabulary compared to the vocabulary of  reformulated questions (used in TNTSPA and Corr. Qsts), since the questions are all formulated using the same template-words. Hence, the reduced variance helps the model understand the questions better. 

\begin{table}
\centering
\begin{tabular}{lcccc}
\toprule
\multicolumn{1}{c}{} & \multicolumn{2}{c}{\textbf{Transformer}} & \multicolumn{2}{c}{\textbf{ConvS2S}}\\
\cmidrule(rl){2-3} \cmidrule(rl){4-5}
\textbf{Dataset} & {BLEU} & {Acc.} & {BLEU} & {Acc.}\\
\midrule
\rowcolor{lavender}
Mon   & 95.86 & 90.55 & 96.35 & 91.66 \\
Mon50 & 96.26 & 91.72 & 95.25 & 88.34 \\
\rowcolor{lavender}
Mon80 & 96.35 & 92.69 & 94.47 & 82.68 \\
\midrule
TNTSPA  & 55.98 & 42.80 & 52.24 & 44.00 \\
\rowcolor{lavender}
Corr. Qsts   & 49.61 & 32.07 & 49.94 & 40.80 \\
Intrm. Qsts & 60.31 & 43.60 & 65.65 & 47.40 \\
\midrule
\rowcolor{lavender}
DBNQA &  64.86 & 46.41 & 67.26 & 45.43 \\
\bottomrule
\end{tabular}
\caption{Performances of baseline architectures on original datasets. \textbf{TNTSPA} is \cite{TNTSPA}'s version of LC-QuAD. \textbf{C. Qsts} designates the LC-QuAD corrected questions and \textbf{Intrm. Qsts} designates the LC-QuAD intermediary questions.}\label{tab:resOG}
\vspace{-1em}
\end{table}







\paragraph{Baseline architectures on tagged datasets.} Table~\ref{tab:resTagged} shows the results of the baseline architectures on tagged datasets. We must not overlook the fact that using tagged data might help the architectures perform better, even without a copy layer. Since the KB elements are encoded as a single symbol, the size of the source and target vocabularies decreases, which usually helps the models perform better. These changes do not make much difference for the Monument datasets since the datasets contain enough examples for the models to learn the KB elements with or without tagging. For the LC-QuAD intermediary questions, we see a clear increase in performance. This is explained by the fact that in the untagged version, the URIs are encoded in the SPARQL query using multiple tokens (\texttt{dbr:Primus\_ attr\_open band attr\_close}), whereas they are encoded as a single token in the NL question and the SPARQL query in the tagged version (\texttt{dbr:Primus\_(band)}). 

For DBNQA, many URIs are quite long and expressed using multiple tokens in the questions. In the untagged version, this means many NL tokens are reused across multiple URI expressions, resulting in a smaller source vocabulary of 99603 tokens. Because there are more unique URIs than unique NL tokens used to represent these URIs in the questions, the tagged version uses a bigger source vocabulary composed of 158014 tokens. However, we see by comparing tables~\ref{tab:resOG} and~\ref{tab:resTagged} that this augmentation of the vocabulary size does not affect the performance of the baseline models.


\begin{table*}[t]
\centering
\begin{tabular}{lcccccccccc}
\toprule

\multicolumn{1}{c}{} & \multicolumn{2}{c}{\textbf{Mon}} & \multicolumn{2}{c}{\textbf{Mon50}} & \multicolumn{2}{c}{\textbf{Mon80}} & \multicolumn{2}{c}{\textbf{Intrm. Qsts}} & \multicolumn{2}{c}{\textbf{DBNQA}}\\

\cmidrule(rl){2-3} \cmidrule(rl){4-5} \cmidrule(rl){6-7} \cmidrule(rl){8-9} \cmidrule(rl){10-11}

\textbf{Architecture} & {BLEU} & {Acc.} & {BLEU} & {Acc.}  & {BLEU} & {Acc.}  & {BLEU} & {Acc.}  & {BLEU} & {Acc.}\\

\midrule
\rowcolor{lavender}
Transformer &  97.02 & 92.81 & 97.41 & 94.41 & 97.80 & 94.86 & 70.29 & 51.93 & 65.63 & 47.75 \\
Transf.-copy & 100 & 100 & 100 & 100 & 100 & 100 & 98.38 & 97.60 & 93.88 & 85.09 \\
\midrule
\rowcolor{lavender}
ConvS2S & 97.82 & 95.26 & 97.71 & 95.13 & 98.14 & 95.96 & 76.62 & 52.93 & 67.57 & 45.22 \\
ConvS2S-copy & 100 & 100 & 100 & 100 & 100 & 100 & 98.35 & 97.40 & 95.40 & 86.87 \\

\bottomrule
\end{tabular}
\caption{Performances of all architectures on tagged datasets}
\label{tab:resTagged}
\end{table*}

\paragraph{Copy-augmented architectures.}  Table~\ref{tab:resTagged} shows the results of our copy-augmented architectures on tagged datasets. We observe a strong increase in performance for LC-QuAD and DBNQA, which is impressive considering the number of different KB elements in the datasets, as well as perfect results on the Monument datasets. However, the most telling results are those obtained on the OOV datasets, reported in Table \ref{tab:resOOV}. The answer accuracy metric is not included because of the high proportion of possible false positives across all OOV datasets. Still, using only the BLEU score, we see that the baseline architectures struggle to handle KB elements they have never seen, which is more representative of the actual capabilities of the models. Similarly, the results on tagged OOV datasets with baseline architectures are still low compared to the results on the original test sets, since tagged data still does not allow the model to adequately handle new KB elements after training. However, on copy-augmented architectures, we observe perfect performances on Monument, representing an increase in performance of about 30 BLEU points compared to its baseline counterpart. On LC-QuAD, 
the increase of about 40 BLEU points shows that the models handle better unknown KB elements using a copy mechanism. 


\setlength{\tabcolsep}{0.2em}
\begin{table*}
\centering
\begin{tabular}{lcccccc}
\toprule
 &  \multicolumn{2}{c}{\textbf{Monument}} & \multicolumn{2}{c}{\textbf{LQ Intrm. Qsts}} & \multicolumn{2}{c}{\textbf{DBNQA}} \\
\cmidrule(rl){2-3} \cmidrule(rl){4-5} \cmidrule(rl){6-7}
\textbf{Dataset} & {Original} & {Tagged} & {Original} & {Tagged} & {Original} & {Tagged} \\
\midrule
\rowcolor{lavender}
Transf      & 60.16 & 65.55 & 51.50 & 56.75 & 40.92 & 41.19 \\
Transf-copy & -     & 100   & -     & 85.68 & -     & 79.82 \\
\midrule
\rowcolor{lavender}
ConvS2S      & 63.88 & 48.31 &  55.85 & 60.98 & 40.62 & 40.66 \\
ConvS2S-copy & -     & 100   & -      & 90.16 & -     & 89.13 \\
\bottomrule
\end{tabular}
\caption{BLEU scores of all the models on the OOV datasets.}\label{tab:resOOV}
\end{table*}

\section{Discussion}

In view of these results, it is clear that, given a working tagging mechanism, the use of a copy-augmented architecture is an excellent advantage for SPARQL NMT architectures as it allows them to handle KB elements not seen in training. Furthermore, comparing the results with and without copy reported in Table \ref{tab:resOOV}, we see a clear improvement in the quality of the translations. 


Another advantage of using a copy-augmented architecture is that it can perform almost as well on small datasets as on larger ones, as demonstrated by the high performances on the LC-QuAD Intermediary Questions and DBNQA. Essentially, the model does not need to learn the correspondences between each expression and the related URI anymore, and it does not need as many examples to learn the templates' formulations since there are not that many. Our work also highlights, as shown by the drastic difference between tables \ref{tab:resOG} and \ref{tab:resOOV}, that baseline models that are reported to have almost perfect performance are, in fact, not as effective outside the test set on which they are evaluated. Even if the BLEU score is a good way to evaluate the quality of the translation,
The use of accuracy and the introduction of OOV datasets helps us understand better a model's actual capabilities.

There is however still room for improvement. 
Some of the limitations of this research lie in the use of template-based entries. In its current state, our copy-augmented architecture depends on questions following specific templates. 
As shown by the results reported in table \ref{tab:resOG}, Seq2Seq models seem quite efficient at learning templates. As we see in Table \ref{tab:resTagged}, the performances increase when the KB elements are encoded in the questions, hinting at the fact that the model is limited by the large amount of KB elements in the dataset rather than the questions' formulations. 
Moving away from template-based datasets would also allow us to determine whether the copy layer helps the model understand the underlying schema of the KB. 

We also need to improve the way OOV datasets are generated to be able to get a representative accuracy metric that is not biased by false positives. To do so, we must ensure most - if not all - queries return a non-empty answer.

Finally, another limitation is that our copy-augmented models depend on tagged questions to reach their top performance.  
\section{Conclusion}
This paper determined that, coupled with a copy-augmented architecture, integrating the KB elements directly in the questions is sufficient for a SPARL NMT model to handle OOV KB elements and to obtain a significant increase in performance. 
These tagged datasets were used to train baseline and copy-augmented versions of the Transformer and the ConvS2S architectures. Using a copy layer, we report perfect performances on the Monument dataset and the generated OOV Monument dataset. For LC-QuAD, we report an increase in BLEU score of 20 points and an increase in answer accuracy of about 40 points. For DBNQA, our results show an increase in BLEU score of 35 points on average, as well as an increase in answer accuracy of 40 points. Our future work will involve the design of a neural tagging model and a joint tagging objective for our Seq2Seq models, as well as the comparison of our copy-augmented models with large pre-trained models and the use of these models as our encoders-decoders. 
Notable models on which to test our methodology include T5 \cite{T5}, BART \cite{BART} and GPT-3 \cite{GPT3}, as well as models that can generate code such as Codex \cite{Codex}. 

\section*{Acknowledgements}
This research has been funded by the NSERC Discovery Grant Program.  

\bibliographystyle{acl_natbib}
\bibliography{MAIN_ARTICLE}

\end{document}